\documentclass{siamart1116}

\numberwithin{theorem}{section}

\usepackage{amsopn}

\usepackage{graphicx}
\usepackage{subfig}
\usepackage{amssymb}
\usepackage{amsmath}
\usepackage{amsbsy}
\usepackage{cleveref}
\usepackage{cancel}
\usepackage{bbm}
\usepackage{cite}
\usepackage[export]{adjustbox}
\usepackage{multirow}

\usepackage{algorithm}
\usepackage[noend]{algpseudocode}
\usepackage{color}

\makeatletter
\def\BState{\State\hskip-\ALG@thistlm}
\makeatother

\def\md{{\bf d}}

\def\mm{{\bf m}}

\def\mr{{\bf r}}

\def\mp{{\bf p}}

\def\mb{{\bf b}}

\def\mrho{{\pmb \rho}}
\def\mbeta{{\pmb \beta}}
\def\rd{{\rm d}}

\def\mrho{{\boldsymbol\rho}}
\title{ Estimate exponential memory decay in hidden markov model and its applications to inference}
\author{Felix X.-F. Ye, Yi-an Ma and Hong Qian}
\begin{document}

\maketitle

\begin{abstract}
Inference in hidden Markov model has been challenging in terms of scalability due to dependencies in the observation data.
In this paper, we utilize the inherent memory decay in hidden Markov models, such that the forward and backward probabilities can be carried out with subsequences, enabling efficient inference over long sequences of observations.
 We formulate this forward filtering process in the setting of the random dynamical system and there exist Lyapunov exponents in the i.i.d random matrices production. 
 And the rate of the memory decay is known as $\lambda_2-\lambda_1$, the gap of the top two Lyapunov exponents almost surely. 
 An efficient and accurate algorithm is proposed to numerically estimate the gap after the soft-max parametrization. The length of subsequences $B$ given the controlled error $\epsilon$ is $B\approx\log(\epsilon)/(\lambda_2-\lambda_1)$.
We theoretically prove the validity of the algorithm and demonstrate the effectiveness with numerical examples. The method developed here can be applied to widely used algorithms, such as mini-batch stochastic gradient method. Moreover, the continuity of Lyapunov spectrum ensures the estimated $B$ could be reused for the nearby parameter during the inference.

\end{abstract}

Hidden Markov model (HMM) and its variants have seen wide applications in time series data analysis.
It is assumed in the model that the observation variable $Y$ probabilistically depends on the latent variables $X$ with {\it emission distribution} $p(y_n | x_n)$ at each time $n$.
The underlying probability of the discrete random variables $X$ follows a Markov chain with {\it transition probability} $p(x_n | x_{n-1})$ \cite{Rabiner1990}.
HMM is the simplest dynamic Bayesian network and has proven a powerful model in many applied fields including 
speech recognition \cite{jelinek1997, Juang1991, Rabiner1990}, 
computational biology \cite{Krogh1994, Krogh2001, Sonnhammer1998}, 
machine translation \cite{Och2000, Och2004}, cryptanalysis \cite{Karlof2003} 
and finance \cite{bhar2006}. Model parameters and hidden variables are inferred for prediction or classification tasks.

Traditionally, model parameters and hidden variables are estimated iteratively for the HMMs through the celebrated Baum-Welch algorithm \cite{murphy2012machine}.
For this maximum likelihood estimation, a forward-backward procedure is used which computes the posterior marginals of all hidden state variables given a sequence of observations.
Later, Bayesian algorithms are also developed through
forward filtering backward sampling algorithm
and variational Bayes method which handles conjugate emission models on the natural parameter space through similar veins as the Baum-Welch algorithm.

In all the aforementioned approaches for inference in HMMs, marginalization over hidden variables is involved.
This step is the crux of the computation burden.
For long observation sequences, this step causes problems of scalability, computation error, and even numerical stability in inference for HMMs \cite{fraser2008hidden, murphy2012machine,Khreich2012}.
Hence an important question is: can one only use part of the data to approximate marginal likelihood over hidden variables of the entire chain, so that stochastic algorithms can be developed with controllable error?

To economize on computational cost at each iteration, we will take advantage of the memory loss property for the filtered state probability. The key idea is that successive blocks of sufficiently long subsequence observations can be considered almost independent of each other.
In this paper, we make use of this memory loss property to approximate the predictive distribution of hidden states $p(x_n | y_{1:n})$ by only using part of the observation sequence $p(x_n | y_{n-B+1:n})$.
This is achieved by formulating $p(x_n | y_{1:n-1})$ as a long sequence of heterogeneous matrices (comprised of emission probabilities and the transition probability) applied successively on an initial probability vector.

However, a critical question that needs to be answered is how long should the subsequence be? Though previous theory exists to quantify the length, the resulting lengths are often longer than the entire sequence which is practically not useful. So one needs to evaluate the rate of memory loss accurately and efficiently to control the length of the subsequence. If we recall the process of calculating filtered state probability, it can be considered as independent and identically distributed random matrix production if we treat observations as random events. It turns out there is a mathematical framework called random dynamical system (RDS) and the long time behavior of random matrices production is described in multiplicative ergodic theorem (MET), also called Oseledec's theorem \cite{rds}. Specially, there exists the Lyapunov spectrum. Previous results showed the rate of memory loss is upper bounded by the gap of the top two Lyapunov exponents, $\lambda_2-\lambda_1$ and is in fact realized almost surely \cite{Atar1997,collet2014}. In particular, the memory loss property requires the Markov chain to be irreducible and aperiodic and the emission distribution to be positive, such that the gap is strictly negative \cite{LeGland2000_1, LeGland2000_2,Atar1997,collet2014}. In this work, we develop an algorithm to accurately and efficiently calculate this gap and the length of subsequence for the given error.

The paper is organized as follows. To make the presentation self contained, in Section \ref{sec-HMM}, we review the basic concepts on hidden Markov models. In Section \ref{sec-forgetting}, we introduce the exponential forgetting of the filtered state probability and review the connection of the forgetting rate and the gap of Lyapunov exponents. In Section \ref{sec-algorithm}, we propose an accurate and efficient algorithm to estimate the forgetting rate and it also provides insight for justification of the gap being the forgetting rate. In Section \ref{sec-example}, we apply this algorithm to estimate the gradient of log-likelihood function efficiently with the help of stochastic gradient descent method. In Section \ref{sec-conclusion}, possible extensions to further speed up the inference are proposed.

 \section{Introduction to HMM}
 \label{sec-HMM}
 Hidden Markov models(HMM) are a class of discrete-time stochastic process $\{X_n, Y_n, n\ge 0\}$: $\{X_n\}$ is a latent discrete valued state sequence generated by a Markov chain, with values taking in the finite set $\{1, 2, \dots, K \}$; $\{Y_n\}$ is corresponding observations generated from distributions determined by the latent states $X_n$. Here it assumes $Y_n$ taking values in $\mathbb{R}^d$, but it can easily extended to discrete states.

We can use the forward algorithm to compute the joint distribution  $p(x_n, y_{1:n})$ by marginalizing over all other state sequences $x_{1:n-1}$.  $Y_n$ is conditionally independent of everything but $X_n$ and $X_n$ is conditionally independent of everything but $X_{n-1}$, i.e, $ p(y_n |x_{1:n}, y_{1:n-1})=p(y_n | x_n)$ and $p(x_n|x_{n-1},y_{1:n-1})=p(x_n|x_{n-1})$. The algorithm takes advantage of the conditional independence rules of HMM to perform the calculation recursively. Using Bayes's rule, we have:
 \begin{align} \label{recursion}
 p(x_n , y_{1:n})&=\sum_{x_{n-1}}p(y_n | x_n)p(x_n|x_{n-1})p(x_{n-1},y_{1:n-1})
 \end{align}
In the above equation, $p(y_n|x_n)$ is referred to as the emission distribution with emission parameters $\{\phi_i\}_{i=1}^K$. Meanwhile, $p(x_n|x_{n-1})$ represents the transition probability of the Markov chain, which is denoted by the transition matrix  $M$. In most cases, we assume $M$ is primitive, i.e, the corresponding Markov chain is irreducible and aperiodic. We denote the parameter of interest as $\theta=\{M, \phi\}$. If the emission distribution is Gaussian distribution, then the emission parameters are the mean $\mu$ and the covariance $\sigma$. Using the notation of vectors and matrix operations, the joint distribution can be represented by a row vector  $\mp_n= p(x_n , y_{1:n}|\theta)$, where its $j$-th component is $p(x_n=j , y_{1:n}|\theta)$.
The forward algorithm can be computed using the following non-homogeneous matrix product.
\begin{align} \label{filtering}
\mp_n=\mp_0MD_1MD_2\dots MD_n
\end{align}
In this equation, $\mp_0$ denotes the initial state distribution $p(x_0)$. The matrix $D_n$ is a diagonal matrix, where its $j$-th entry is given by $D_{jj}(y_n)=p(y_n | x_n=j, \phi_j)$. This represents the emission distribution when the current state is $j$. Moreover, if one considers the observation $y_n$ as random events, then $D(y_n)$ are random matrices that are  independently sampled at each step. If one starts with invariant distribution of the Markov chain initially, $\pi$, then these matrices are sampled in i.i.d manner with probability density distribution
 \begin{equation}\label{pdf}
 f(y)=\sum_j \pi_j p(y | x_n=j, \phi_j)\end{equation}
If the initial distribution is not $\pi$, after couple time steps, the distribution follows the invariant distribution and one can assume these matrices are sampled in i.i.d manner anyway. Now it is turned into a product of random matrices problem and these  diagonal matrices randomly rescale the columns of $M$. $\mp_n$ is called forward probability.

If we normalize the vector $\mp_n$, it obtains the {\it filtered state probability}, $\mrho_n=p(x_n |y_{1:n}, \theta)$, which is not the invariant distribution of the Markov chain,
\begin{equation} \label{filtering-proj}
\mrho_n=\frac{p(x_n , y_{1:n}|\theta)}{p(y_{1:n}|\theta)}=\frac{\mp_n}{\mp_n\cdot \mathbbm{1}}
\end{equation}
This process is called filtering. The normalization constant $(\mp_n\cdot \mathbbm{1})$ gives the total probability for observing the given sequence up to step $n$  irrespective of the final states, which is also called marginal likelihood $p(y_{1:n}|\theta)$. Not only this process ensures the numerical stability of random matrices production, but also $\mrho_n$ provides the scaled probability vector of being each state at step $n$.
Note the probability vector $\mrho_n$ lives in a simplex, $S^{K-1}$, which is also called projective space in dynamical system, or space of measure in probability theory. Instead, the joint probability $\mp_n$ is in $\mathbb{R}^{K+}$.

Another joint probability column vector $\mb_i=p(y_{i+1:n}|x_{i},\theta)$ is the probability of observing all future events starting with a particular state $x_{i}$. It can be computed by the backward algorithm similarly and it is called backward probability. We begin with $\mb_n=\mathbbm{1}$, and it gives
\begin{align}
\mb_i=MD_{i+1}\dots MD_n\mathbbm{1}
\end{align}
It is again a product of random matrices. One can similarly renormalize the backward probability vector for better numerical stability, $\mbeta_i=\mb_i/(\mb_i\cdot \mathbbm{1})$ such that $\mbeta_i\propto p(y_{i+1:n}|x_{i},\theta)$. In fact, with forward and backward probability, we can calculate the probability $p(x_i | y_{1:n},\theta)\propto \mrho_i^T\circ\mbeta_i$ which is the Hadamard product of two vectors. In fact,
the entry for the highest entry of this probability vector can give rough idea which latent state at step $i$ lies.

\section{Exponential Forgetting}
\label{sec-forgetting}
Heuristically, in this very long heterogenous matrix multiplication (\ref{filtering}), one observes that the final vector is irrelative to the initial vector and almost determined by the last several matrices multiplications, up to a normalization constant. As a matter of fact, if one is not interested in the precise value of the final vector, the subsequence of matrices with length $B$ are sufficient to approximate the vector.
In more mathematical precise writings: Start any two different initial state probability vector $\mp_0$ and $\mp'_0$ and after applying exactly the same sequence of matrices, they generate two sequence of filtered state probability $\mrho_n$ and $\mrho'_n$. The distance of two sequence goes to 0 asymptotically almost surely, i.e,
\begin{align}\label{sync}
\lim_{n\rightarrow +\infty}\|\mrho_n-\mrho'_n\|=0\ a.s.
\end{align}
This phenomenon is called  {\it loss of memory of HMM}. 

{\bf Example:} In the figure \ref{fig-sync},  Markov chain has three state, emission distribution is a one-dimensional Gaussian on each state and the parameter $\phi$ is

$$M=\begin{bmatrix} 0.005 & 0.99 & 0.005 \\0.01 &0.03 & 0.96 \\ 0.95 & 0.005 & 0.045 \end{bmatrix}, \mu=[0, 0.5, -0.5], \sigma=[1, 1, 1]$$
 Starting with every point in the simplex as initial conditions, we apply these points by the same sequence of random matrices. One observes that the triangle consisting all points starts to shrink along $n$ and after 40 steps, the triangle is contained within a small circle with radius $\epsilon$. As $n$ goes to $+\infty$, it will synchronize into a random fixed point, since it is sequence dependent. That implies if one allows error of $\epsilon$, it may only requires the last 40 matrices which is irrelevant with the initial condition. So it significantly simplifies computational complexity.

 If the diagonal matrices $D_i$ are homogenous, it degenerates to the corollary of Perron-Frobenius theorem for primitive matrices.
Now natural questions to arise are: what are conditions for such phenomenon and under these conditions, what are the rate of convergence. This rate in fact answers the critical question that how long the length $B$ should be for a given $\epsilon$. More questions about the rate are how to estimate the rate numerically or even analytically and does the rate continuously depends on the parameter $\theta$.

In fact, the sufficient conditions for this phenomenon are given in
Le Gland {\it et al} \cite{LeGland2000_1,LeGland2000_2},

\begin{theorem}
\label{LeGland}
If Markov transition matrix is primitive and the emission distribution $p(y_n|x_n)$ is strictly positive, then for any $\mp_0, \mp'_0\in S^{K-1}$, there exists a strictly negative $-c$
\begin{equation}
\limsup_{n\rightarrow +\infty} \frac{1}{n}\log\|\mrho_n-\mrho'_n\|\le -c, \ \ \text{almost\ surely}
\end{equation}
\end{theorem}

The theorem implies the filtered state probability forget almost surely their initial conditions exponentially fast and the rate is at least $c$.  So the phenomenon of {\it loss of memory of HMM} is also called {\it exponential forgetting of prediction filter}. 
The techniques they used are Hilbert metric and Birkhoff contraction coefficient $\tau(M)$, which are extensively applied in non-negative matrix theory \cite{seneta1981, hartfiel2002}. Definitions of both terms are included in the appendix \ref{app-Birkhoff} and it also showed that $\tau(M)<1$ for positive matrix $M$ which is a sub-class of primitive matrix.
It is a bit surprising that eigenvalues of each matrix in the heterogenous matrix production have little to do with this asymptotic behavior. In particular, one can construct a matrix sequence that spectrum radius of each is uniformly less 1, but the product doesn't even converge to 0. It is because the spectral radius doesn't process sub-multiplicity property, on the other hand, this Birkhoff contraction coefficient does. Moreover, $\tau(M)=0$ if and only if each row of $M$ is a scalar multiple of the first row, which is also called weak ergodicity. At last, this coefficient is invariant with rescaling rows and columns of matrix. From these three properties, one immediately concludes when $M$ is positive, the heterogeneous matrix production in (\ref{filtering}) has the weak ergodicity and the exponential forgetting of the prediction filter follows with convergence rate $\log\tau(M) $. To further relax the positive matrix to primitive matrix, the approach is rather technical. 

On the other hand, the long time behavior of random matrices production is well studied in {\it multiplicative ergodic theorem} (MET) through Lyapunov exponent. It is the heart of a field called {\it Random Dynamical System} (RDS) \cite{rds}. Lyapunov exponent is like the generalization of logarithm of absolute value of eigenvalues in the terms of random matrices production. Atar {\it et al} \cite{Atar1997} and Collet {\it et al} \cite{collet2014} gave the exact convergence rate by Lyapunov exponents,

\begin{theorem}
\label{Atar}
\begin{equation}
\limsup_{n\rightarrow +\infty} \frac{1}{n}\log\|\mrho_n-\mrho'_n\| =\lambda_2-\lambda_1,  \ \ \text{almost\ surely}
\end{equation}
\end{theorem}

So the convergence rate is upper bounded by the gap between the first two Lyapunov exponents of the products of random matrices in (\ref{filtering}) and in fact realized for almost all realizations. Furthermore, they showed this gap is strictly negative when the transition matrix is primitive and the emission distribution is positive. Then it recovered Le Gland's results.
There is a nice connection between two results: Peres \cite{Peres1992}  proved the gap of the first two Lyapunov exponents, $\lambda_2-\lambda_1$ in i.i.d random matrices production is upper bounded by $\log \tau(M)$. So for positive matrices, these two results connect with each other naturally.

Random dynamical system, as an extension of the theory of non-autonomous dynamical system, has different setup from stochastic process and is somewhat inaccessible to a nonspecialist. Here we will present this theory in the setting of product of random matrices intuitively. The rigorous definition is included in appendix \ref{app-rds}.

We will describe an i.i.d RDS for the sake of convenience and one can extend easily to independent but not identical RDS. Results presented in this paper can be extended to ergodic case.
The state space is $\mathbb{R}^{K+}$ and the family of matrices $\Gamma$ is all the possible diagonal matrices $D$. We would like to study the dynamics of $\mp_n$ in (\ref{filtering}). Although $\mp_n$ itself has probability meaning, here we are merely treating it as $K$ dimensional random variable.  Initially, starting from an initial condition $\mp_0$, a diagonal matrix $D_1=D(y_1)$ is chosen according to the probability density distribution $f(y)$ in (\ref{pdf}). Then the system moves to the state $\mp_1=\mp_0 MD_1$ in step 1. Again, independently of previous maps, another matrix $D_2=D(y_2)$ is chosen according to the same probability density function and the system moves to the state $\mp_2=\mp_1 MD_2$. The procedure repeats. The random variable $\mp_n$ is now constructed by means of multiplication of independent random matrices.

The asymptotic limit of the rate of growth for the product of independent random matrices, $\lim_{n\rightarrow +\infty}\frac{1}{n}\log\frac{\|\mp_n\|}{\|\mp_0\|}$, is as been studied started at the beginning of the 60's. It has great relevance for development of the ergodic theory of dynamical system. Furstenberg and Kesten \cite{Furstenberg1960, Furstenberg1963} showed
\begin{theorem}
\begin{equation}
\lambda_1=\lim_{n\rightarrow +\infty}\frac{1}{n}\log\frac{\|\mp_n\|}{\|\mp_0\|},  \ \ \text{almost\ surely}
\end{equation}
this limit $\lambda_1$ exists almost surely, moreover, it is a nonrandom quantity and independent of the choice of metric.
\end{theorem}
 It is considered as the extension of strong law of large number to i.i.d random matrices \cite{PRM}. This limit is called {\it maximum Lyapunov exponent}. It is rather surprising result since the order of sequence seems not much important even for non-commutative matrix multiplication. However, the Furstenberg-Kesten theorem neglects the finer structure given by the lower growth rates, other than the maximum Lyapunov exponent. Later Oseledets \cite{MET} showed there exists Lyapunov spectrum $\Lambda$, like eigenvalue spectrum, from the multiplicative ergodic theorem (MET). Similarly Lyapunov spectrum doesn't depend on the choice of sequence almost surely and thus it is a global property for this random matrix multiplication. For a given initial vector, such set of sequences that gives different asymptotic limit of growth rate has zero measure.
Analog with eigenvector, it also has Lyapunov vector which describes characteristic expanding and contracting directions, but it depends on the particular ergodic sequence. The statement of the theorem is included in appendix \ref{app-rds}.

The filtered state probability $\mrho_n$ is projected onto the simplex $S^{K-1}$ in (\ref{filtering-proj}) and the dynamics of it will be an induced RDS. There is a nice theorem connecting Lyapunov spectrum of both RDS. \cite{rds}
\begin{theorem}
 \label{splitting}
 Lyapunov spectrum of the induced RDS is that of the corresponding RDS subtracts the maximum Lyapunov exponent, i.e, $\Lambda'=\Lambda-\lambda_1$.
 \end{theorem}
Specifically, when the condition in theorem \ref{LeGland} is fulfilled, then maximum Lyapunov exponent of the induced RDS, $\lambda'_1=0$ with multiplicity 1 and the next one is $\lambda'_2=\lambda_2-\lambda_1$ which is what we desire to estimate.

In the framework of RDS, the exponential forgetting property defined above is equivalent with  {\it synchronization by noise}.
Synchronization is the phenomenon when trajectories of random dynamical systems subjected to the same randomness, but starting from different initial vectors converge in time to a single (random) solution almost surely, like in eq (\ref{sync}). So these trajectories are not independent. Synchronization has been widely discovered as a relevant property in modeling of external noises. In neurosciences, one observes
this synchronization by common noise as a reliable response of one single neural oscillator
on a repeatedly applied external pre-recorded input, which may be seen as a dynamical system
driven by the same noise path but different initial conditions. \cite{Lin2013, Guillaume2013}

However, we note not every RDS processes this property. Specifically, Newman \cite{newman} showed the necessary and sufficient conditions for stable synchronization in continuous state RDS. Crudely speaking, in order to see synchronization, one needs two ingredients: local contraction (negative maximum Lyapunov exponents) so that nearby points approach each other; along with a global irreducibility condition. In discrete state RDS, conditions for synchronization are discussed as well \cite{Felix}. In HMM, the global irreducibility holds since the transition matrix $M$ is primitive and the local contraction is guaranteed by this gap $\lambda_2-\lambda_1$. It recovers the results previously obtained. Much intuitive picture will be presented in the next section.
 So the 2-norm of difference for two nearby trajectories has the following behavior,
\begin{align}
\|\mrho_n-\mrho'_n\|\le C\exp\Big((\lambda_2-\lambda_1)n\Big)\|\mp_0-\mp'_0\|
\end{align}
where $C$ is some constant. If one would like to have the error within the radius of $\epsilon$, then the length of the subsequence should be $B\approx\frac{\ln(\epsilon)}{\lambda_2-\lambda_1}$.
However, from the previous literature \cite{LeGland2000_1,  LeGland2000_2, Atar1997}, the explicit analytical estimate of the gap $\lambda_2-\lambda_1$ for a given parameter is either too loose or still difficult to find. So the numerical algorithm of efficient estimation is on demand.

\section{Algorithm}
\label{sec-algorithm}
In fact, one could sample two sequences of $\mrho$ and $\mrho'$ with the same matrices sequence and monitor the maximum length needed to achieve $\epsilon$ error. However, it suffers numerical instability and lack of robustness, such that some rare cases could deviate the estimate.
Or one use QR decomposition directly to find the Lyapunov spectrum for (\ref{filtering}) which takes about $O(K^3)$ order of multiplications per each iteration. But what needed is merely the second largest one instead of the whole spectrum.  Then it is possible to have a more efficient algorithm and  may provide some new insight that why this gap governs the exponential forgetting rate.
In realistic scenario, one may possibly access the forward probability $\mp_n$ or the filtered state probability $\mrho_n$, or at least some portion of them. We would like to take advantage of these information without redoing this time-consuming filtering process.

\begin{figure}
\begin{tabular}{cc}
{\bf (a)} \includegraphics[valign=t,scale=0.28]{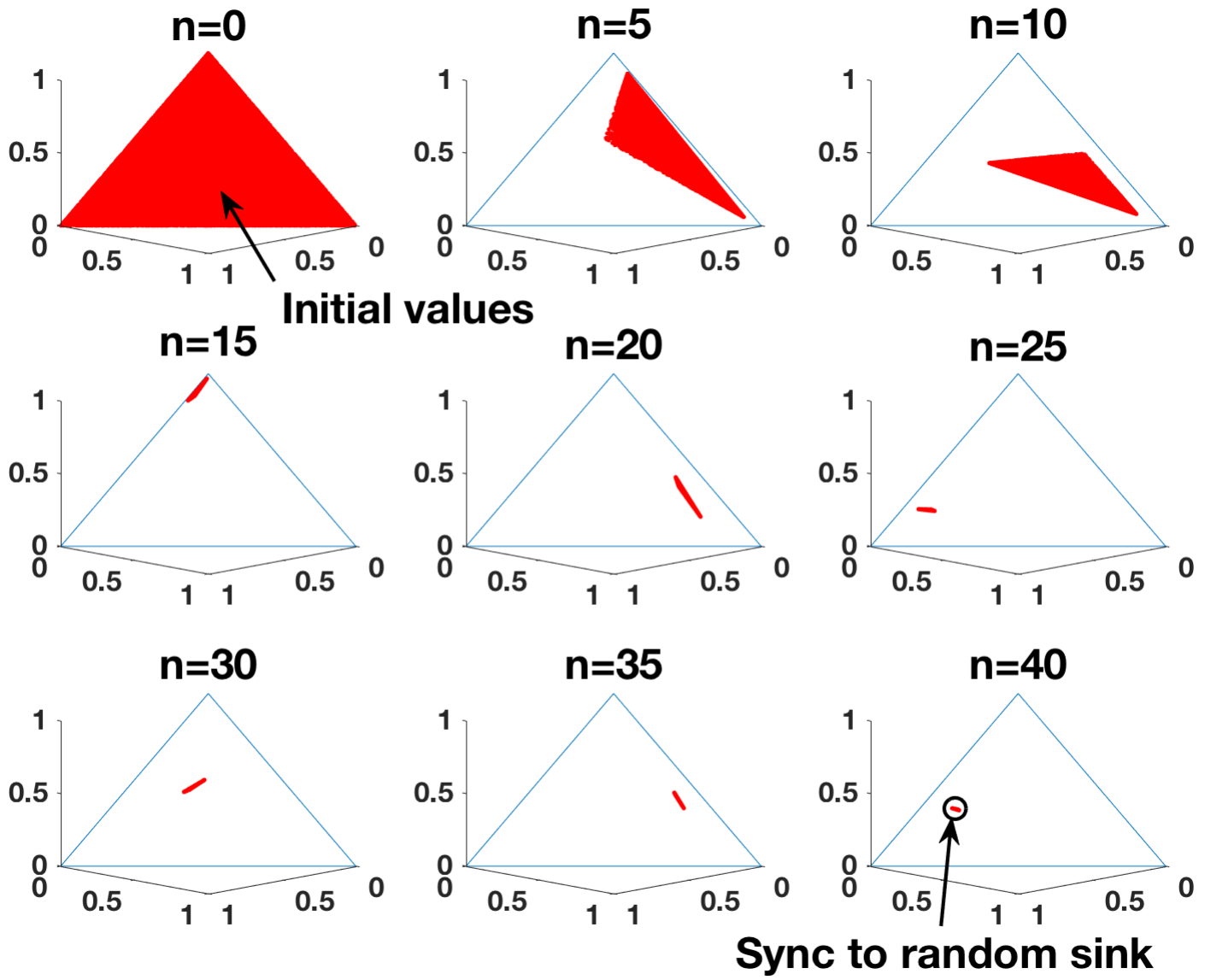} &
{\bf (b)} \includegraphics[valign=t,scale=0.3]{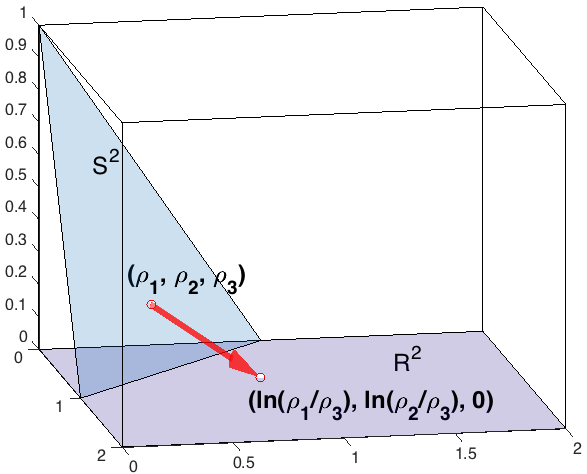}
\end{tabular}
\caption{(a)  Starting with every point in the simplex, apply the same sequence of random matrices, and the triangle is contained within a small circle with radius $\epsilon$ after 40 steps. (b) Diagram of the projection from a point in the simplex $S^2$ to $\mathbb{R}^2$.}
\label{fig-sync}
\end{figure}

 If $\mrho=[a_1, a_2, \dots, a_K]$, define a projection $\Pi$ from simplex $S^{K-1}$ to $\mathbb{R}^{K-1}$ as the log ratio relative to the last component. The projection is illustrated for the example in figure \ref{fig-sync}.
\begin{align} \label{proj}
\Pi: \mrho=\begin{bmatrix}a_1, a_2, \dots, a_K \end{bmatrix} \rightarrow \mr=\Big[\underbrace{\log(\frac{a_1}{a_K})}_{r_1}, \underbrace{\log(\frac{a_2}{a_K})}_{r_2}, \dots, \underbrace{\log(\frac{a_{K-1}}{a_K})}_{r_{K-1}}, 0\Big]
\end{align}
Denote $r_K=0$ as convention, such that $\mr$ is embedded in $\mathbb{R}^{K}$. Since $M$ is primitive, $a_K$ cannot be 0 except for at most $K$ initial steps. In the mean time, $\mrho$ will be in the interior of the simplex. Such projection from compact space to non-constraint space, illustrated in figure \ref{fig-sync}, is relatively common in numerical optimization which is called soft-max parametrization. It directly implies the constraint condition $\sum_i a_i=1$ and $a_i>0$.  The inverse of the projection, $\Pi^{-1}$ is
\begin{equation}
\Pi^{-1}: \mr=\begin{bmatrix}r_1, r_2, \dots, r_{K-1}, 0\end{bmatrix}\rightarrow \mrho=\Big[\underbrace{\frac{\exp(r_1)}{\sum_i \exp(r_i)}}_{a_1},\underbrace{\frac{\exp(r_2)}{\sum_i \exp(r_i)}}_{a_2}, \dots, \underbrace{\frac{\exp(r_K)}{\sum_i \exp(r_i)}}_{a_K}\Big]
\end{equation}
The index of the summation is from 1 to $K$. This projection naturally defines an induced RDS for the dynamics of $\mr$. Furthermore,

\begin{theorem}
 \label{project}
If the coordinate transformation is bijective, and both its derivative and inverse exist, then the Lyapunov spectrum remains invariant under such a transformation.
\end{theorem}
Then the projection preserves the Lyapunov spectrum.
It also means the synchronization with the variable $\mr$ implies the synchronization with $\mrho$ and vice versa. Heuristically understanding, $\lambda'_1=0$ is due to the constraint condition and after the parametrization, the condition is inherited in the last component $r_K=0$. If we only study the dynamics for the first $K-1$ unconstrained coordinates, it removes this particular Lyapunov exponent of the induced RDS but keeps the rest of the spectrum the same. Now the maximum Lyapunov exponent is the desired difference $\lambda_2-\lambda_1$.

 In addition, the dynamics of $\mr$ has the following nice property. The random map $G_\md$ for $\mr$ has the form as
\begin{align} \label{random-map}
\mr_{n+1}&=G_{\md_n}(\mr_n)=\md_n+F(\mr_n)
\end{align}
It is composed with random translation $\md$ and deterministic map $F(\mr)$.
Each component of the map $F$ is explicitly given as
\begin{align}\label{deterministic-map}
F_i(\mr)&=\ln \Big(\frac{\sum_{j=1}^{K}\exp(r_j)M_{ji}}{\sum_{j=1}^{K}\exp(r_j)M_{jK}}\Big),\ 1\le i \le K-1
\end{align}
If we denote $\mm_i$ as the $i$-th column of the transition matrix $M$ and $\exp(\mr)$ as component-wise exponent, eq (\ref{deterministic-map}) can be rewritten by inner product form
\begin{align}
F_i(\mr)=\ln \Big(\frac{\exp(\mr)\cdot\mm_i}{\exp(\mr)\cdot\mm_K}\Big),\ 1\le i \le K-1
\end{align}
The random translation is similarly defined as the log ratio of diagonal of $D$ relative to the last component,
$\md_n=\begin{bmatrix}\ln \frac{p(y_n|x_n=1)}{p(y_n|x_n=K)}, \dots, \ln \frac{p(y_n|x_n=K-1)}{p(y_n|x_n=K)}\end{bmatrix}$. Since the emission distribution is positive, the log ratio is well defined.
The random map is the translation of the deterministic smooth map $F$ by the i.i.d random variable $\md_n$ and $F$ is solely dependent on the transition matrix $M$.
It is even more interesting to notice the Jacobian of this random map is independent with $\md$, it is $J(\mr)=\nabla F(\mr)$ since the random translation will not affect the local contraction or expansion.

The $(K-1)$-by-$(K-1)$ Jacobian matrix $J(\mr)$ can be explicitly expressed as follows,
\begin{eqnarray}
J_{ij}(\mr)=\frac{\exp(\mr_j)M_{ji}}{\exp(\mr)\cdot \mm_i}-\frac{\exp(\mr_j)M_{jK}}{\exp(\mr)\cdot\mm_K},\  1\le i,j\le K-1
\end{eqnarray}

Then we will have the corollary following by Theorem \ref{splitting} and Theorem \ref{project}
\begin{corollary}
\begin{align}
\lambda_2-\lambda_1=\limsup_{n\rightarrow +\infty}\frac{1}{n}\log \|J(\mr_n)J(\mr_{n-1})\cdots J(\mr_1))\|
\end{align}
\end{corollary}

Now the maximum Lyapunov exponent $\lambda_2-\lambda_1$ is approximated by the finite time Lyapunov exponent \cite{Strogatz, Barreira}.
Instead of using QR decomposition, the maximum Lyapunov exponent can be estimated by averaging finite time approximations which is much faster and easier to implement. One can start with a unit test vector, apply these Jacobian matrices sequentially and renormalize the vector at each step. Averaging all these renormalization constants along the timeline will give the approximation of maximum Lyapunov exponent. It is not a concern that all vectors are alignment along the direction of maximal expansion because we are not interested in finer structure of the spectrum. The order of multiplication needed is $O(K^2)$ per each iteration which is faster than QR decomposition. More importantly, if one already has partial data set of the filtered state probability, they can be projected to $\mr$ and estimate the Lyapunov exponent directly without further information on observed sequences.

\begin{figure}\centering
\begin{tabular}{cc}
{\bf (a)} \includegraphics[valign=t,scale=0.31]{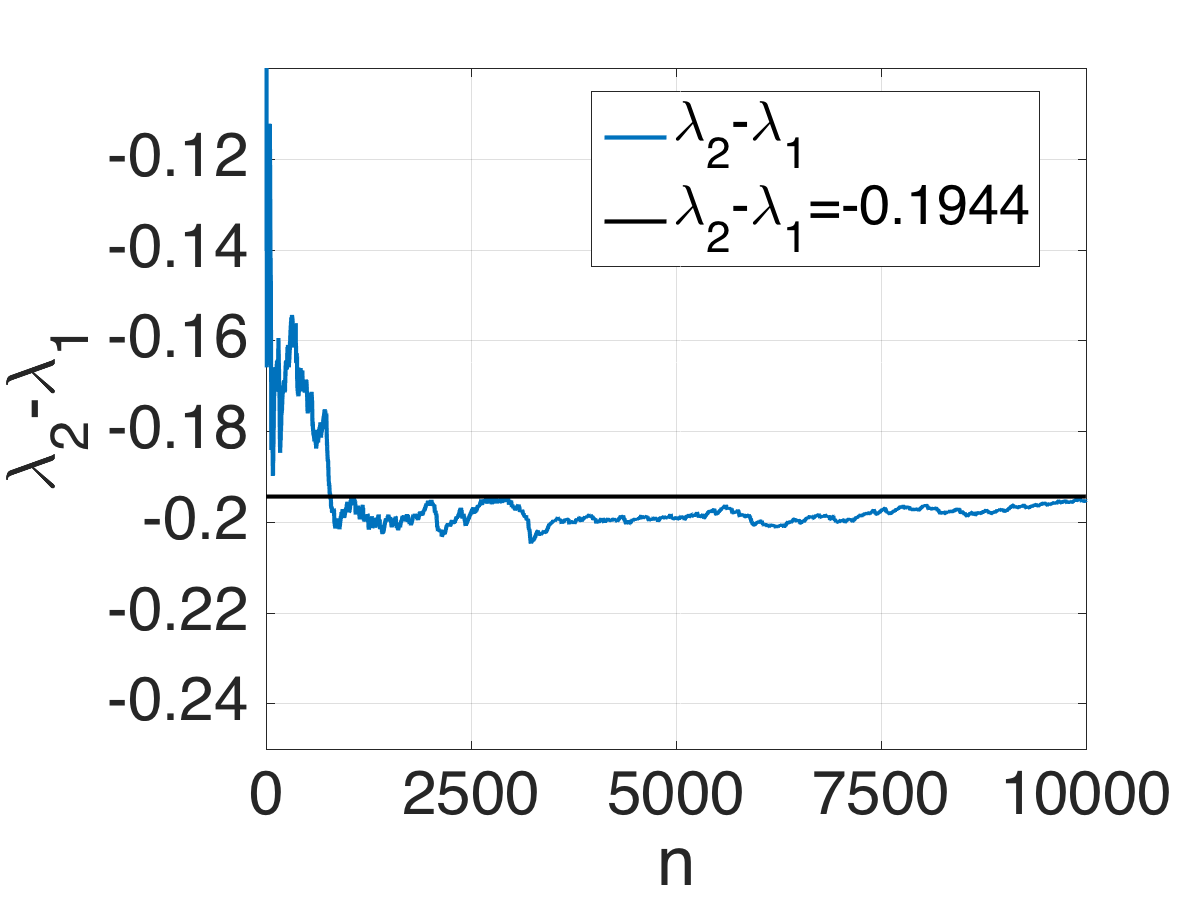} &
{\bf (b)} \includegraphics[valign=t,scale=0.31]{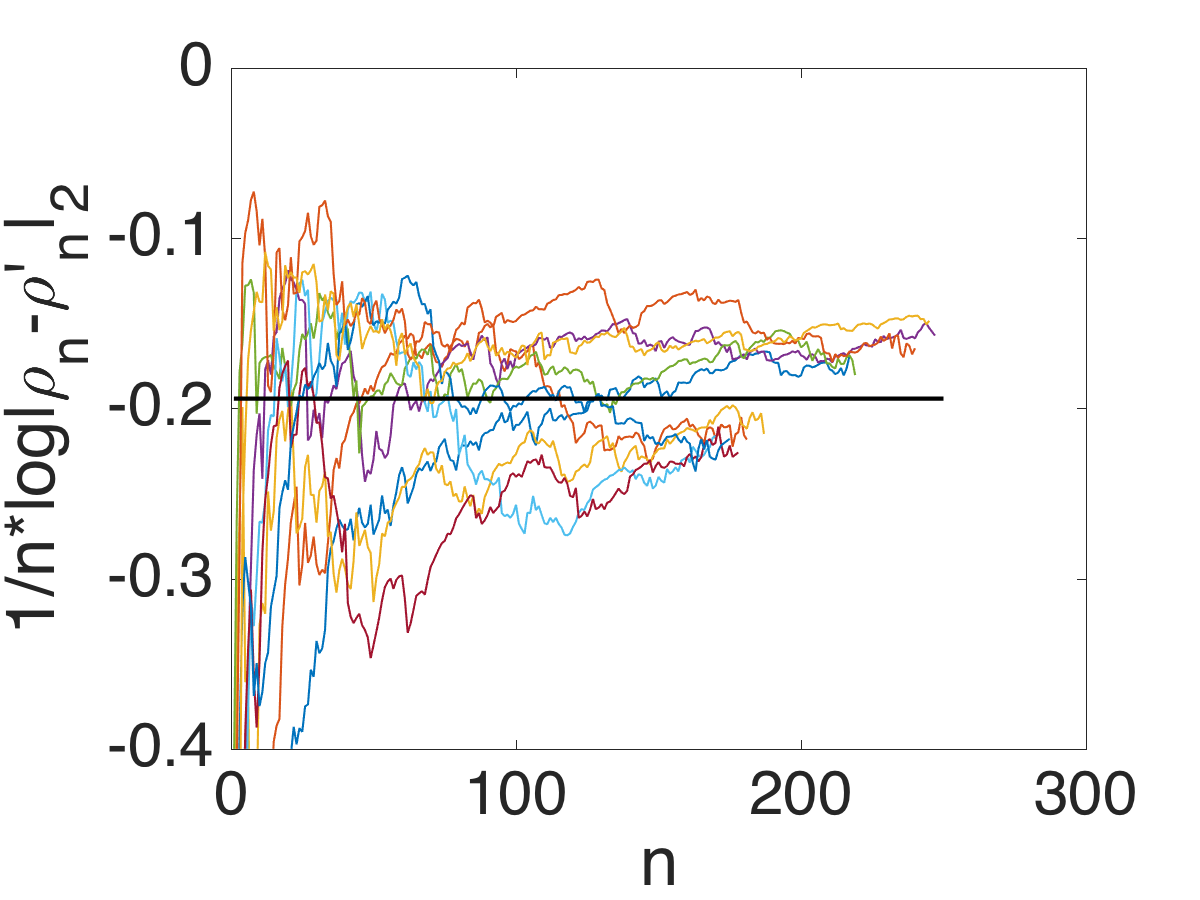}
\end{tabular}
\begin{tabular}{c}
\ \ \ \ \ \ \ \ \ \ \ {\bf (c)} \includegraphics[valign=t,scale=0.33]{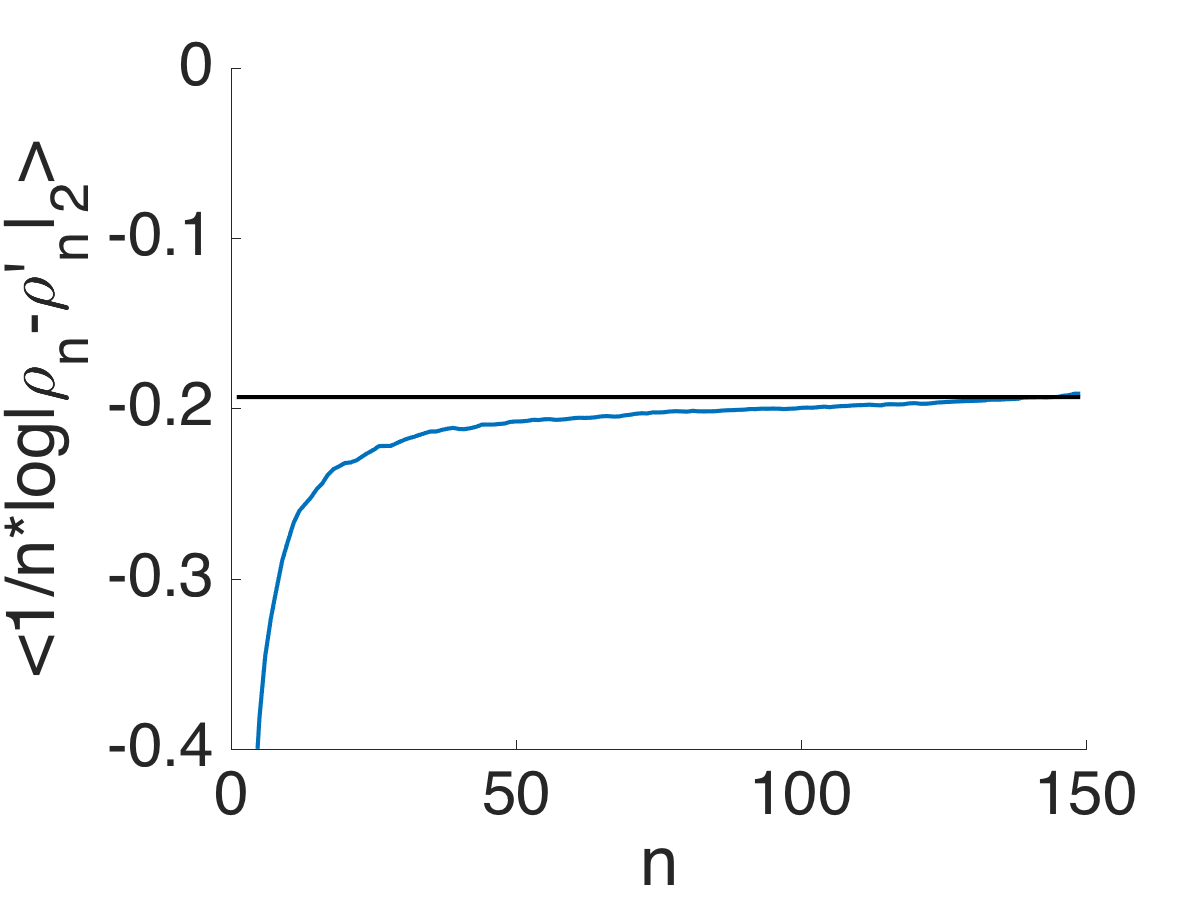}
\end{tabular}
\caption{(a) We use algorithm \ref{alg-1} to estimate the gap of Lyapunov exponent with the observation sequence with length of 10000.
(b) We sample 10 independent sequences for $\frac{1}{n}\log \|\mrho_n-\mrho'_n\|_2$ and compare with the theoretical limit (black line). (c) We average 500 independent sample sequences and compare with the theoretical limit (black line).  }
\label{fig-lyap}
\end{figure}
The maximum Lyapunov exponent for this induced random map $\lambda_2-\lambda_1$ characterizes the rate of separation of infinitesimally close trajectories in $\mathbb{R}^{K-1}$. If two vectors $\mr'$ and $\mr$ are close enough, one could use their difference to approximate the 2-norm of the difference of $\mrho'$ and $\mrho$, $\|\mrho'-\mrho\|_2\le  \frac{1}{4}\|\mr'-\mr\|_2$. Then the rate of separation for $\mrho$ in fact is upper bounded by the gap $\lambda_2-\lambda_1$, which is the estimation of exponential forgetting rate. This algorithm provides some new insight for the analytical justification for the gap.

We apply this algorithm to approximate the gap of Lyapunov exponent in the previous example. In the figure \ref{fig-lyap}, the estimated gap is $\lambda_{\text{max}}=-0.1944$ with data of 10000 and the length $B$ needed for $\epsilon=10^{-15}$ is about 178. On the other hand, one starts with two different initial conditions $\mp_0$ and $\mp'_0$ and applies the same sequences of random matrices to obtain $\mrho$ and $\mrho'$ after normalization. We plot $\frac{1}{n}\log \|\mrho_n-\mrho'_n\|_2$ along $n$ for ten independent sequences and they roughly converge to the theoretical limit $-0.1944$. However, as $n$ increases, $\|\mrho_n-\mrho'_n\|_2$ reaches the machine epsilon and becomes numerically unstable, such that some sequences are cut off beyond $n=150$. So we are not able to visualize the strong convergence directly. If we average 500 sample sequences, then we can clearly visualize the convergence in mean. With the uniform integrability, convergence in mean is granted by the strong convergence.

Right now, it seems one needs to estimate the gap for each parameter $\theta$. One related result is if matrices are nonsingular and maximum Lyapunov exponents are simple, then it depends continuously on the probability \cite{Peres1992}. Bocker and Viana \cite{viana2017} showed Lyapunov spectrum depend continuously on matrices and probability for 2-dimensional case, as far as all probabilities are positive. Moreover, a few of Avila's deepest results with Eskin and Viana \cite{avila2023continuity}, extend the statement to arbitrary dimension. The book \cite{viana2014} gives a nice introduction on this most recent approach. The direct consequence for this result is it doesn't need to estimate the gap every time and it is safe to reuse the previous estimation for couple steps in parameter inference.
The pseudocode of estimating the length $B$ is given in Algorithm \ref{alg-1}.

\begin{algorithm}
\caption{Estimate the length $B$}\label{alg-1}
\begin{algorithmic}[1]
\State $a\gets0$, $\text{initialize} \ \mp_0\  $ and $e$,
\State $\textbf{for}\  i=0,1,\dots,N_{iter}, \textbf{Do}$
\State \ \ \ \ $\mp_{i+1}\gets\mp_{i}MD_{i+1}$, $D_{i+1}$ is given in (\ref{filtering}),
\State \ \ \ \ $\mrho_{i+1}\gets\mp_{i+1}/(\mp_{i+1}\cdot \mathbbm{1}) $, update $\mr_{i+1}$ according to (\ref{proj}),
\State \ \ \ \ $e\gets J(\mr_{i+1})e$, $a\gets a+\log\|e\|$, $e\gets e/\|e\|$.
\State $\textbf{end for}$
\State $\lambda\gets a/N_{iter}, B\gets\log(\epsilon)/\lambda$.
\end{algorithmic}
\end{algorithm}

\section{Applications}
\label{sec-example}
\subsection{Statistical Inference}
Traditionally, EM, variational inference or MCMC are used to perform inference over $\theta$. These algorithms have found widespread use in statistics and machine learning \cite{murphy2012machine, fraser2008hidden}. However, it is a computational challenge in terms of scalability and numerical stability, to marginalize all hidden state variables given a long sequence of observations.
There are many other gradient based algorithms to obtain the maximum likelihood estimator(MLE) or maximum a posteriori (MAP), for instance, stochastic gradient descent method.  We must be able to efficiently estimate the gradient of the log-likelihood function or log-posterior function, $ \ln p(\theta |y_{1:n})$. The likelihood function $p(y_{1:n}|\theta)$ is written as 
$p(y_{1:n}|\theta)=\sum_j p(x_n=j, y_{1:n}|\theta)= \mp_n\mathbbm{1}=\mp_i\mb_i$ for $0\le i\le n$.

 With the prior function $p(\theta)$, the gradient is written as
\begin{align} \nonumber
\frac{\partial \ln p(\theta |y_{1:n})}{\partial \theta_i}&= \frac{\partial \ln p(y_{1:n}|\theta)}{\partial \theta_i}+\frac{\partial \ln p(\theta)}{\partial \theta_i} \\ \nonumber
& = \sum_{j=1}^n\frac{\mp_{j-1}\frac{\partial MD_j}{\partial \theta_i} \mb_j}{\mp_{j-1}MD_j  \mb_j}  +\frac{\partial \ln p(\theta)}{\partial \theta_i}\\ \label{original-gradient}
&=\sum_{j=1}^n\frac{\mrho_{j-1}\frac{\partial MD_j}{\partial \theta_i} \mbeta_j}{\mrho_{j-1}MD_j \mbeta_j}  +\frac{\partial \ln p(\theta)}{\partial \theta_i}
\end{align}

The complexity of matrix multiplication needed to calculate one component of the gradient is $O(n)$ and the space needed is also $O(n)$. So it is prohibitively expensive to compute directly in space and time when $n$ is very large. Moreover, this direct computation is not numerically stable since the numerator and denominator are usually extremely small in such massive matrix multiplication.
In fact, there are various algorithms to reduce the complexity, including the following mini-batch gradient descent method, which employs noisy estimates
of the gradient based on minibatch of data \cite{Yian2017, Foti2014, Fox2019}.

First, instead of summing over all index $j$ from 1 to $n$, uniformly sample a subset of summand $S$ with cardinality $s$ at each step and use the following estimator for the direction of the full gradient. Here we assume the prior distribution $p(\theta)$ as uniform for the sake of simplicity,
 \begin{align} \label{rewrite-gradient}
 \frac{\partial \ln \widetilde{p}(\theta |y_{1:n})}{\partial \theta_i}= \frac{n}{s}\sum_{j\in S}\frac{\mrho_{j-1} \frac{\partial MD_j}{\partial \theta_i}\mbeta_{j} }{\mrho_{j-1}MD_j \mbeta_{j}}
 \end{align}
Then we expect $\mathbb{E}( \frac{\partial  \ln\widetilde{p}(\theta |y_{1:n})}{\partial \theta_i})= \frac{\partial \ln p(\theta |y_{1:n})}{\partial \theta_i}$. This is typically referred to mini-batch gradient descent based techniques and it is very effective in the case of large-scale problems.

Second, instead of computing normalized forward and backward probability $\mrho_j$ and $\mbeta_j$ recursively,  we introduce a buffer of length $B$ on left and right ends of the subsequence of random matrices and both vectors are estimated by this much shorter subsequence,

 \begin{align} \label{filtering-approx}
& \widetilde{\mp}_{LB}=\mp_0 MD_{j-B}\dots MD_{j-1},\ \ \widetilde{ \mrho}_{j-1}= \widetilde{\mp}_{LB}/(\widetilde{\mp}_{LB}\cdot \mathbbm{1}) \\
&  \widetilde{\mb}_{RB}=MD_{j+1}\dots MD_{j+B}\mathbbm{1}, \ \ \widetilde{\mbeta}_j=  \widetilde{\mb}_{RB}/( \widetilde{\mb}_{RB}\cdot \mathbbm{1})
 \end{align}
The reason to use the same buffer length for forward and backward probability is that  Lyapunov spectrums for forward algorithm and backward algorithm are exactly the same. Therefore, the gap of the top two Lyapunov exponents are the same and the buffer of length is the same. 

\begin{theorem}
Let $\theta$ be an invertible ergodic measure-preserving transformation of a probability space $(\Omega, \mathbb{P})$, where $\mathbb{P}=Q^\mathbb{Z}$ is the Bernoulli measure. Let $A: \Omega\to M(d, \mathbb{R})$ satisfy $\int\log \|A_\omega^{\pm 1}\|\rd \mathbb{P}<\infty$. Let $P_n(\omega)$ be the induced linear cocycle, $P_n(\omega)=A(\theta(n-1)\omega)\dots A(\omega)$. Then the Lyapunov spectrums of these two linear cocycles 
\begin{align}
\lambda:=\lim_{n\rightarrow +\infty}\frac{1}{n}\log \| {\bf u} P_n(\omega) \|, \ \lambda':=\lim_{n\rightarrow +\infty}\frac{1}{n}\log \|P_n(\omega){\bf v} \|
\end{align}
are equal $\mathbb{P}$-a.s. 
\end{theorem}
The proof is straight forward. The left multiplication and the right multiplication will not change the Lyapunov spectrum but the subspaces $U_i(\omega)$ will be different for two cocycles. 

So the gradient (\ref{rewrite-gradient}) is approximated as
\begin{align} \label{final-gradient}
 \frac{\partial \ln \widetilde{p}(\theta |y_{1:n})}{\partial \theta_i}=\frac{n}{s}\sum_{j\in S}\frac{\widetilde{\mrho}_{j-1}\frac{\partial MD_j}{\partial \theta_i}\widetilde{\mbeta}_j}{\widetilde{\mrho}_{j-1}MD_j \widetilde{\mbeta}_j}
\end{align}
Note that (\ref{final-gradient}), the matrix multiplication required is $O(2Bs)$ after using the buffer and the space needed is $O(s)$. When $2Bs\ll n$, this results in significant computational speedups over the full batch inference algorithm. This techniques is exactly due to the memory decaying property and the buffer length $B$ is calculated in Algorithm \ref{alg-1}. In order to be consistent with this technique, we can uniformly sample the subset of summand in the domain $[B+1, n-B-1]$. Moreover, one can enforce no overlap among the sampled subsequences. In pseudocode, the algorithm is presented in Algorithm \ref{alg-2}.

\begin{algorithm}
\caption{Mini-batched based inference for HMM}\label{alg-2}
\begin{algorithmic}[1]
\State initialize the parameter $\theta=\{M, \phi\}$ and learning rate $\eta$
\State $\textbf{for}\  i=0,1,\dots,N_{iter}, \textbf{Do}$
\State \ \ \ \  Periodically estimate the buffer length $B$ according to Algorithm \ref{alg-1}
\State \ \ \ \  i.i.d sample $s$ integers in the subset $[B+1, n-B-1]$ with uniform distribution.
\State \ \ \ \  Calculate $\frac{\partial \ln \widetilde{p}(\theta |y_{1:n})}{\partial \theta}$ according to (\ref{final-gradient}).
\State \ \ \ \  $d=\frac{\partial \ln \widetilde{p}(\theta |y_{1:n})}{\partial \theta}/\|\frac{\partial \ln \widetilde{p}(\theta |y_{1:n})}{\partial \theta}\|$
\State \ \ \ \ $\theta\gets \theta+\eta d$

\State $\textbf{end for}$
\end{algorithmic}
\end{algorithm}

This memory decay property not only can be taken advantage of in mini-batched gradient descent based inference on MLE or MAP, but also be used in stochastic gradient-MCMC \cite{Yian2017, completesample, Fox2019}, stochastic variational inference \cite{Foti2014}, stochastic EM and online learning \cite{Khreich2012}. There are many more algorithms could be built based on this fundamental property in HMM. No matter what mini-batched based algorithm is used, it is important to estimate the buffer length efficiently and accurately.

\subsection{Synthetic Example}
\begin{figure}\centering
\includegraphics[scale=0.3]{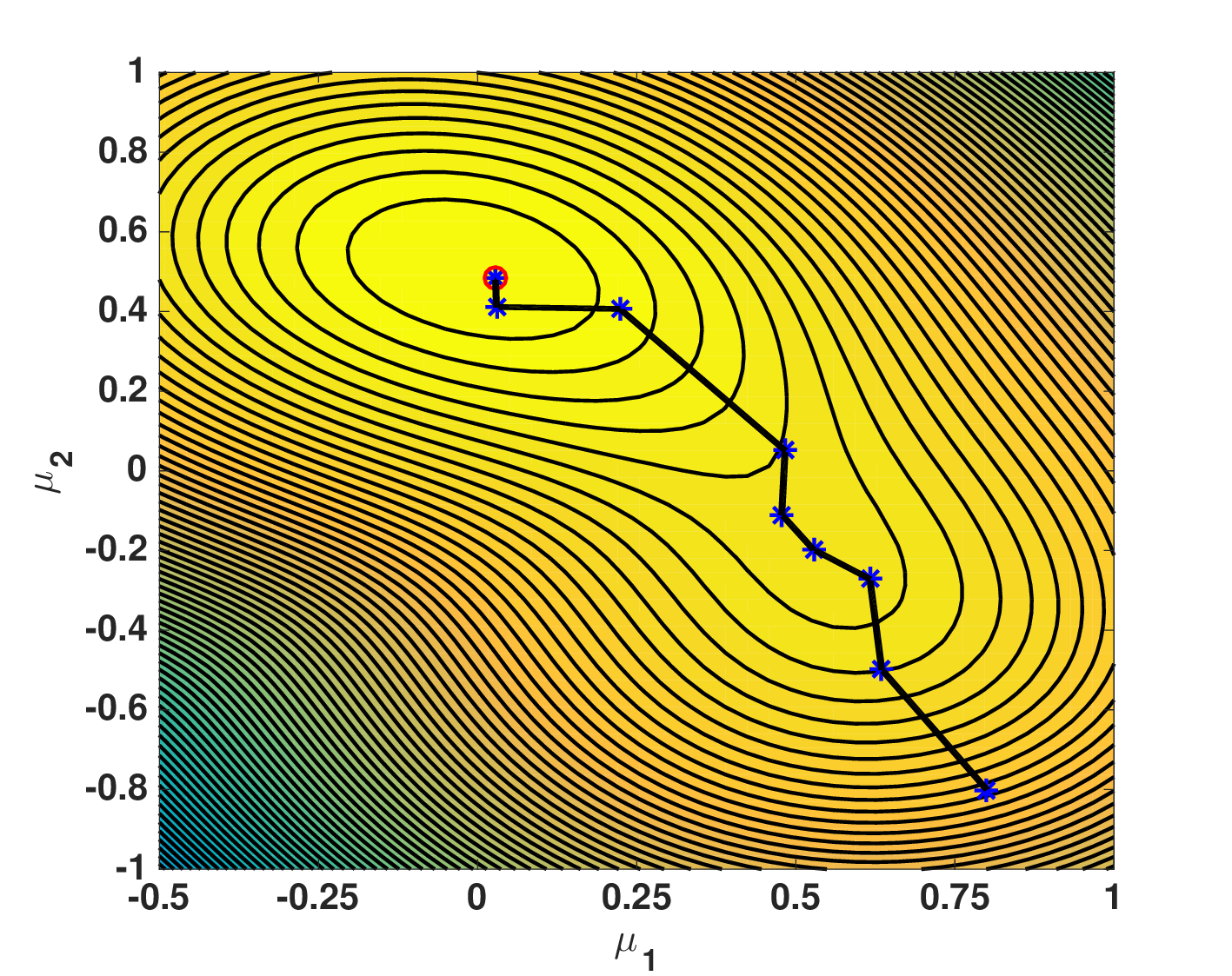}
\includegraphics[scale=0.3]{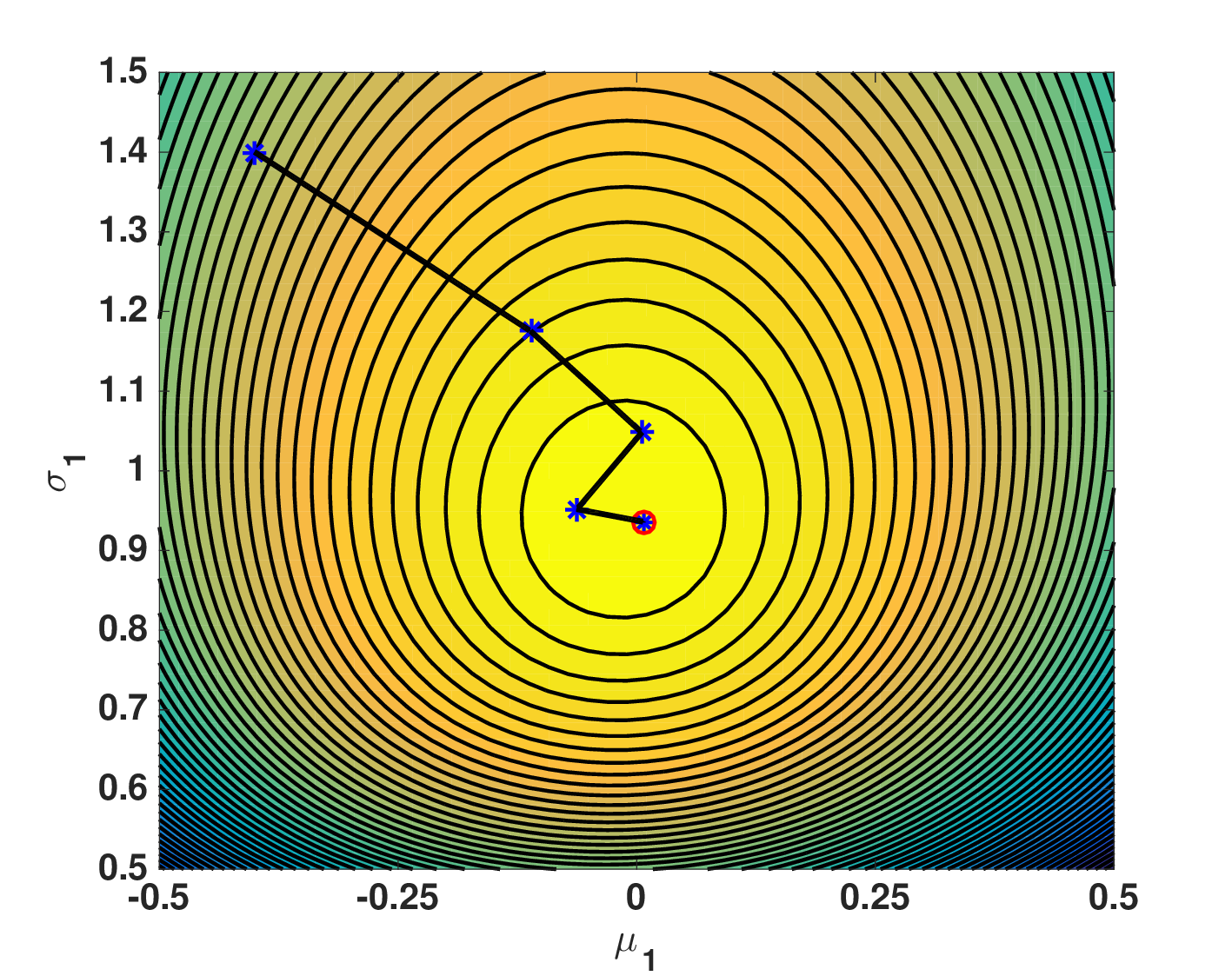}
\caption{Apply the Algorithm \ref{alg-2} to Example 1. The background is the contour plot of the log-likelihood function. In the left figure, $\mu_1$ and $\mu_2$ are unknown, the algorithm converges to $(0.03,0.48)$ starting from $(0.8,-0.8)$. In the right figure, $\mu_1$ and $\sigma_1$ are unknown, the algorithm converges to $(0.01, 0.94)$ starting from $(-0.4,1.4)$. }
\label{MLE}
\end{figure}

In order to demonstrate the algorithm, we sampled a long observation sequence with length $L=10^7$ and the parameter given in example 1. We assume we know all parameters except $\mu_1$ and $\mu_2$. In the algorithm, we use the same left and right buffer length $B=200$ and sample size $s=100$.
The learning rate $\eta$ starts with 0.05 and decays with the rate of 0.95 along the steps to prevent oscillations. After 25 steps, the algorithm will restart with the latest parameter, until the difference of parameter from the previous restart is within the threshold, which in this case is 0.02.  From the figure \ref{MLE}, the parameter reaches the desire MLE $(0.03, 0.48)$ after 8 starts from the initial guess $(0.8, -0.8)$. Note from the contour plot of the log-likelihood function, there is a region around $(0.5, 0)$ which is the flip of the mean, is very flat and our algorithm is able to escape it due to the stochastic nature. Another remark is the matrix multiplication needed is about $8*10^6$ which is less than the length observation sequence $L=10^7$. It implies the algorithm has reached MLE even before the filtering procedure is finished in single iteration in EM or gradient descent algorithms. So it significantly speed up the inference procedure.

In our recent work, we leveraged Algorithm \ref{alg-2}, which demonstrated its efficacy in larger systems  in \cite{Yian2017}. We subsequently employ two synthetic examples presented in \cite{Yian2017} as our second and third examples. Comprehensive details of these examples can be found in Appendix \ref{app-syn}.
To optimize the performance of our Algorithm \ref{alg-2}, it's important to estimate the buffer parameter 
$B$, especially when explicit values aren't readily available in their paper. As suggested in \cite{Yian2017}, the buffer 
$B$ is set as 
 \begin{align}
 B = \frac{1}{\lambda_2-\lambda_1}\log\left(\frac{10^{-3}}{2}\right)
 \end{align}
 where $10^{-3}$ is the error tolerance and 2 is the maximum initial error for the probability vectors. 
 
 For the second example, characterized as the \textit{diagonally dominant} (DD) model, the Markov chain that exhibits heavily self-transitions and has identifiable emissions. Through our algorithm, we estimate $\lambda$ to be approximately -0.0157, which subsequently results in  $B$ being approximately 485.
Conversely, the third example, termed the \textit{reversed cycle} (RC), encapsulates strong transitions between two cycles over three states, each operating in the opposing direction. We estimate
$\lambda$ to be approximately -0.0619, which subsequently results in 
$B$ being approximately 123.

\section{Extension and Conclusion}
\label{sec-conclusion}
Although traditionally the EM algorithm has monotonic convergence, ensures parameters constraints implicitly and generally easier to be implemented, the convergence of EM can be very slow in terms of time for each iteration and total iteration steps. Our method can significantly reduce the time for each iteration since we only utilize part of the data by harnessing the memory loss property, but the steps needed is still comparable with EM algorithm. In fact, one could extend our idea to efficiently estimate the Hessian matrix such that much faster quadratic convergence can be achieved with second-order method. Moreover, one can calculate observed fisher information which is the negative Hessian matrix of the log-likelihood evaluated at MLE. It will give curvature information at MLE and help to decide which MLE may be better without calculating the likelihood explicitly.
Another natural extension of our method is to discrete state Kalman filter, which is continuous time version of HMM. Similar exponential forgetting property and the rate being the gap of top two Lyapunov exponents are discussed in \cite{Atar1997} for the Wonham filter, but the proof is harder and evolves different techniques, in particular, the naive time discretization may be challenging since one needs to justify the change order of the limit. Further extension to continuous state Kalman filter is not feasible at this stage because our method is based on finite dimension random matrix theory. Last but not the least, it is tempering in realistic scenario to use the same buffer length for the left and right subsequences. The numerical simulation indicates the gap of Lyapunov exponents for both forward and backward probabilities for this particular example are close and one would speculate the gaps are the same. To formulate this problem more explicitly, it is equivalent to find the connection of Lyapunov spectrum for i.i.d random matrix production and for their transpose. One result is the Birkhoff contraction coefficient for the matrix is the same as its transpose. So if the Markov transition matrix is positive matrix, then both gaps are bounded by $\log \tau(M)$. But the explicit connection still remains as an open question.

In the era of big data, data analysis methods in machine learning and statistics, such as hidden Markov models, play a central role in industry and science. The growth of the web and improvements in data collection technology in science have lead to a rapid increase in the magnitude and complexity of these analysis tasks. This growth is driving the need for the scalable algorithms that can handle the ``Big Data''. However, we don't need that the whole massive data, instead small portion of data could serve as good as the original. One successful examples are mini-batched based algorithms. Despite the simple chain-based
dependence structure, apply such algorithms in HMM are not obvious, since subsequences are not mutually independent.
 However, with the data set being abundant, we are able to harness the exponential memory decay in filtered state probability and appropriately choose the length of the subchain with the controlled error, to design mini-batched based algorithms.
 We proposed an efficient algorithm to accurately calculate the gap of the top two Lyapunov exponents, which helps to estimate the length of the subchain. We also prove the validity of the algorithm theoretically and verified it by numerical simulations.
 In the example, we also proposed the mini-batched gradient descent algorithm for MLE of log-likelihood function and it significantly reduces the computation cost.

\bibliography{master}
\bibliographystyle{plain}

\appendix

\section{Birkhoff contraction coefficient}
\label{app-Birkhoff}
We will introduce Hilbert metric and Birkhoff contraction coefficient on positive matrices, especially on positive stochastic matrices \cite{seneta1981,hartfiel2002}.

Let $x$ and $y$ be positive vectors in $\mathbb{R}^n$, the Hilbert metric is defined as $d(x,y)=\ln \frac{\max_i x_i/y_i}{\min_j x_j/y_j}$. But Hilbert metric is not a metric in $\mathbb{R}^n$ since one could check when $x=cy$ for some constant $c$, $d(x,y)=0$. Actually, for each positive probability vector in the interior of the simplex $S^{K-1}$, $d$ determines a metric on them.

The advantage of Hilbert metric for the positive stochastic matrix $M$ is one can show for two different positive probability row vector, $x$ and $y$, the distance between $x$ and $y$ under $M$ monotonically decreases, $d(xM, yM)< d(x,y)$. This is not guaranteed for other metrics due to the possible non-normal behavior of the matrix. The Birkhoff contraction coefficient $\tau(M)$ is defined as the supreme of the contraction ratio under the matrix $M$,
\begin{equation}
\tau(M)=\sup\frac{d(xM, yM)}{d(x, y)}
\end{equation}
This coefficient indicates how much $x$ and $y$ are drown together at least after multiplying by $M$. Actually, there is an explicit formula for computing $\tau(M)$ in terms of the entries of $M$. Define $\phi(M)$ as
\begin{equation}
\phi(M)=\min_{p, q, r, s}\frac{M_{pq}M_{rs}}{M_{rq}M_{ps}}
\end{equation}
The term $\frac{M_{pq}M_{rs}}{M_{rq}M_{ps}}$ is cross ratios of all $2\times 2$ sub matrices of $M$ and $\phi(M)$ is the minimum amount of them. If there is a row with both zero and positive elements, $\phi(M)=0$.  The formula for $\tau(M)$ is
\begin{equation}
\tau(M)=\frac{1-\sqrt{\phi(M)}}{1+\sqrt{\phi(M)}}
\end{equation}
As expected, for positive stochastic matrix $M$, $\tau(M)<1$.

\section{Random dynamical system} \label{app-rds}
In this section, we review some important definitions and concepts in random dynamical system(RDS) in terms of this HMM problem. This material can be found in standard textbooks \cite{rds}. It is presented for the convenience of the readers.

 Random dynamical system defines on the metric dynamical system $\Big(\Omega, \mathcal{F}, \mathbb{P}, \theta\Big)$. $\Omega$ is the set of all possible one sided infinitely long sequence of invertible matrices, and $\mathcal{F}$ is the $\sigma$-algebra of $\Omega$.
$$ \Omega=\{\omega: (A_1A_2 \dots ) |A_i\in \mathbb{R}^{d\times d}\}$$
The probability measure for given first $k$ values is
$$\mathbb{P}(A_1A_2 \dots A_k)=f(A_1) f(A_2)\dots f(A_k) \rd\lambda_1\rd\lambda_2\dots\rd \lambda_k $$
where $f(\cdot)$ is the probability density function given in (\ref{pdf}). It is clear that the random variable at each time step are i.i.d.
The $\theta$ is a left shift operator on $\omega$,
$$\theta(A_1A_2  \dots )=(A_2 A_3\dots)$$

The state space is $\mathbb{R}^{d}$ equipped with $\sigma$-algebra $\mathcal{B}$,
and the mapping $\phi(n, \omega)$ defined on the state space is $\phi(n, \omega)=A(\theta(n-1)\omega)\circ A(\theta(n-2)\omega) \cdots \circ A(\omega)$, where $A$ is the matrix valued function and $A(\omega)$ takes the first matrix in the sequence. In particular, $\phi(n, \omega)=A_nA_{n-1}\cdots A_1$. It is the product of i.i.d random matrices. This mapping has cocycle property, i.e, it satisfies
$\phi(0,\omega)=id$ for all $\omega\in \Omega$.
and $\phi(s+n, \omega)=\phi(s, \theta(n)\omega)\circ \phi(n, \omega) $ for all $n,s \in \mathbb{Z}$, $\omega\in \Omega$.

 A skew product of $\theta$ and $\phi(\cdot, \omega)$ is a measurable transformation $S(n)$: $\Omega\times \mathbb{R}^{d}\rightarrow \Omega\times \mathbb{R}^{d}$, defined by
$$S(n): (\omega, v)\rightarrow \Big(\theta(n)\omega, \phi(n, \omega)v \Big)$$

This RDS induces a Markov process on $\mathbb{R}^{d}$ and assume this Markov process has an invariant measure $\nu$.
There is a simple one-to-one correspondence between invariant measure of RDS and induced Markov process: A product measure $\mu=\mathbb{P}\times \nu$ is $S$-invariant, moreover, if $\nu$ is ergodic, then $\mu$ is ergodic.


The multiplicative ergodic theorem states as follows,
\begin{theorem}
Let $\theta$ be an ergodic measure preserving transformation of $(\Omega, \mathbb{P})$, Let $A: \Omega\rightarrow M_{d\times d}(\mathbb{R})$ be a matrix-valued function with $\int \log \|A(\omega)\|\rd \mathbb{P}(\omega)<\infty$. Then there exist $\infty >\lambda_1>\lambda_2 \dots \lambda_k\ge -\infty$; $m_1, \dots, m_k\in \mathbb{N} $ satisfying $m_1+\dots+m_k=d$ and a measurable family of subspaces $F_1(\omega), F_2(\omega), \dots, F_k(\omega)$ such that
\begin{enumerate}
\item filtration:  $\mathbb{R}^{d}=F_1\supset F_2(\omega)\supset \dots F_{k-1}(\omega)\supset F_{k}=\{0\}$.

\item dimension: dim $F_i(\omega)=m_i+\dots+m_k$; for a.e. $\omega$.

\item equivariance: $A(\omega)F_i(\omega)\supset F_i(\theta(\omega))$ for a.e. $\omega$.

\item growth: If $v\in F_i(\omega)\backslash F_{i+1}(\omega)$, then $\frac{1}{n}\log\|\phi(n,\omega)v\|\rightarrow \lambda_i$ for a.e. $\omega$.
\end{enumerate}
\end{theorem}
The Lyapunov spectrum $\Lambda$ of RDS is defined as $\frac{1}{n}\log\|\phi(n,\omega)v\|$. The multiplicative ergodic theorem states for almost all $\omega$ and each non-zero vector $\mr$, the Lyapunov spectrum $\lambda$ exists, depends on $v$ up to $k$ different values but independent of the choice of the metric.

\section{Details on synthetic examples 2 and 3} \label{app-syn}
The second example, \textit{diagonally dominant} (DD) consists of a Markov chain that heavily self-transitions. Most subchains in a minibatch thus contain redundant information with observations generated from the same latent
state. Although transitions are rarely observed, the emission means are set to be distinct so that this
example is likelihood-dominated and highly identifiable. The transition matrix and emission parameters used for this experiment were:

\[ A_{DD} = \left( \begin{array}{cccccccc}
.999 & .001 & 0 & 0 & 0 & 0 & 0 & 0 \\
0 & .999 & .001 & 0 & 0 & 0 & 0 & 0 \\
0 & 0 & .999 & .001 & 0 & 0 & 0 & 0  \\
0 & 0 & 0 & .999 & .001 & 0 & 0 & 0 \\
0 & 0 & 0 & 0 & .999 & .001 & 0 & 0 \\
0 & 0 & 0 & 0 & 0 & .999 & .001 & 0 \\
0 & 0 & 0 & 0 & 0 & 0 & .999 & .001 \\
.001 & 0 & 0 & 0 & 0 & 0 & 0 & .999
\end{array} \right) . \]

\[\boldsymbol{\mu}_{DD} = \left\{
(0,20); (20,0); (-30,-30); (30,-30); (-20,0); (0,-20); (30,30); (-30,30);
\right\}\]
and $\Sigma_{DD} = I$ for all states.

The third example we consider contains two \textit{reversed cycles} (RC): the Markov chain strongly transitions
from states $1 \rightarrow 2 \rightarrow 3 \rightarrow 1$ and $ 5 \rightarrow 7 \rightarrow 6 \rightarrow 5$ with a small probability of transiting
between cycles via bridge states $4$ and $8$.  The emission means for the two cycles are very similar
but occur in reverse order with respect to the transitions.
The emission variance is larger, making states $1$ and $5$, $2$ and $6$, $3$ and $7$ indiscernible by themselves.
Transition information in observing long
enough dynamics is thus crucial to identify between states $1, 2, 3$ and $5, 6,
7$.

The transition matrix and emission parameters were:

\[ A_{RC} = \left( \begin{array}{cccccccc}
.01 & 0 & .85 & 0 & 0 & 0 & 0 & 1 \\
.99 & .01 & 0 & 0 & 0 & 0 & 0 & 0  \\
0  & .99 & 0 & 0 & 0 & 0 & 0 & 0 \\
0 & 0 & .15 & 0 & 0 & 0 & 0 & 0 \\
0 & 0 & 0 & 1 & .01 & 0 & .85  & 0 \\
0 & 0 & 0 & 0 & .99 & .01 & 0 & 0 \\
0 & 0 & 0 & 0 & 0 & .99 & 0 & 0 \\
0 & 0 & 0 & 0 & 0 & 0 & .15 & 0
\end{array}\right) . \]

\begin{equation*}
\boldsymbol{\mu} = \left\{ (-50,0); (30,-30); (30,30); (-100,-10); (40,-40); (-65,0); (40, 40); (100,10)  \right\},
\end{equation*}
and $\Sigma_{RC} = 20*I$ for all states.

\end{document}